\documentclass[letterpaper, 10 pt, conference]{ieeeconf}
\IEEEoverridecommandlockouts   
\usepackage{graphics}
\usepackage{multirow}
\usepackage{graphicx}
\usepackage{adjustbox}
\usepackage{balance}
\usepackage{hyperref}
\usepackage{amsmath}
\usepackage{breqn}
\usepackage{siunitx}
\usepackage[ruled,vlined]{algorithm2e}
\usepackage{hhline}
\usepackage{todonotes}

\usepackage{todonotes}
\newcommand{\revision}[1]{\textcolor{black}{#1}}

\title{\LARGE \bf  Optimization-based Trajectory Planning for Tethered Aerial Robots*
}

\author{S. Mart\'inez-Rozas$^{1}$, D. Alejo$^{1}$, F. Caballero$^{1}$ and L. Merino$^{1}$
\thanks{*This work has been supported by the Spanish Ministry of Science, Innovation and Universities (COMCISE RTI2018-100847-B-C22, MCIU/AEI/FEDER, UE).}
\thanks{$^{1}$S. Mart\'inez-Rozas, D. Alejo, F. Caballero and L. Merino are with Service Robotics Laboratory, Universidad Pablo de Olavide, Seville, Spain. Email: {\tt\small simon.martinez$@$uantof.cl}, {\tt\small daletei$@$upo.es}, {\tt\small fcaballero$@$us.es}, {\tt\small lmercab$@$upo.es}}%
}

\begin{document}
\maketitle
\thispagestyle{empty}
\pagestyle{empty}

\begin{abstract}
This paper presents a non-linear optimization method for trajectory planning of tethered aerial robots. Particularly, the paper addresses the planning problem of an unmanned aerial vehicle (UAV) linked to an unmanned ground vehicle (UGV) by means of a tether. The result is a collision-free trajectory for UAV and tether, assuming the UGV position is static. The optimizer takes into account constraints related to the UAV, UGV and tether positions, obstacles and temporal aspects of the motion such as limited robot velocities and accelerations, and finally the tether state, which is not required to be tense. The problem is formulated in a weighted multi-objective optimization framework. 
Results from simulated scenarios demonstrate that the approach is able to generate obstacle-free and smooth trajectories for the UAV and tether.
\end{abstract}

\section{Introduction}
\label{sec:introduction}
During the last decade, UAVs and UGVs have been employed with success in a great variety of tasks potentially dangerous for humans, such as disaster management, abandoned mines mapping and inspection, or even sewer inspection \cite{jfralejo, 7784277}. The great manoeuvrability of UAVs enables accessing remote areas that would be impossible to reach with UGVs. However, the limited flight time of small UAVs (few tens of minutes) reduces the applicability of this technology. To increase their endurance, tethered UAVs fixed to a base station have been proposed, but the use of a tether usually limits \revision{the UAV range to a great extent}. On the other hand, UGVs have a much larger autonomy, but they may not be capable of reaching certain areas of interest.

\begin{figure}[!t]
    \centering
    \includegraphics[width=0.9\linewidth]{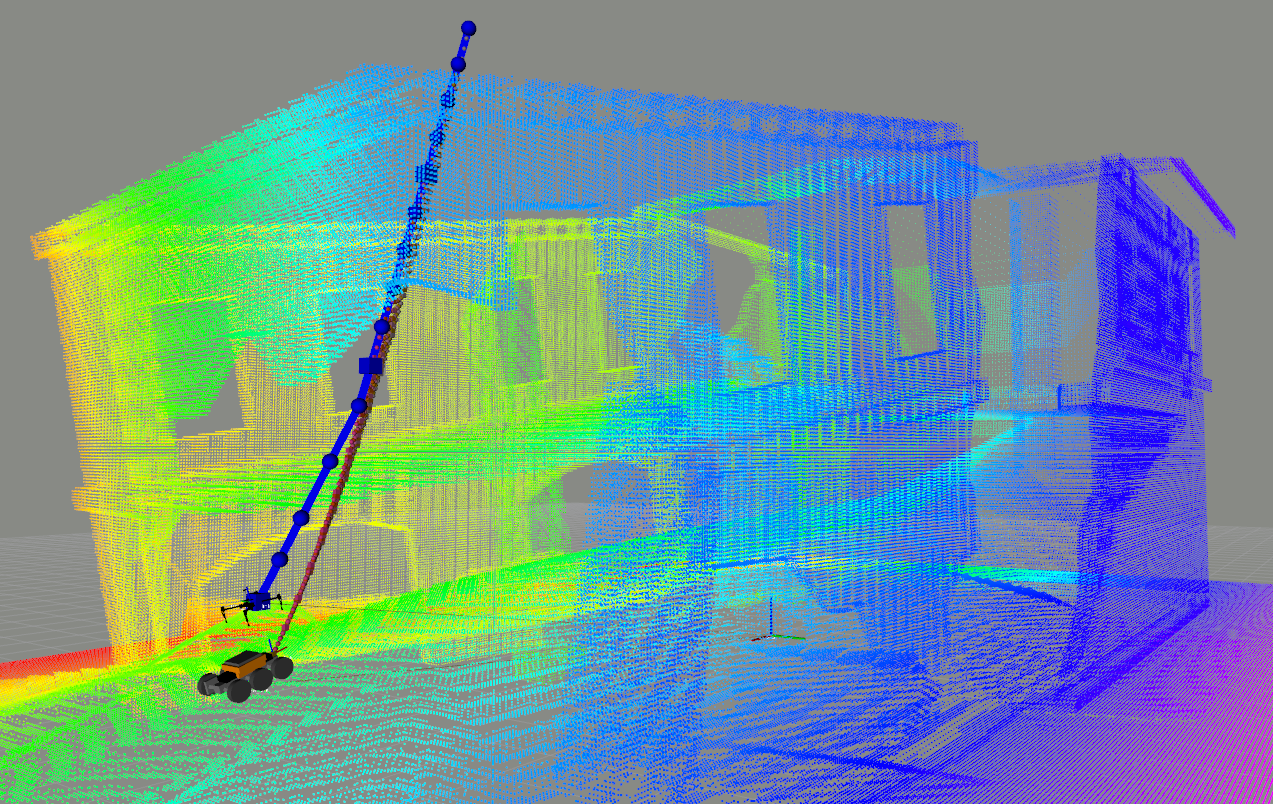}
    \caption{Example of the proposed robotic system. The UAV is linked by a power tether to a UGV. The blue line denotes the computed trajectory considering the tether constraint.}
    \label{fig:intro}
\end{figure}

The idea to take the best of the ground and aerial robots was proposed in the late 90s through the definition of marsupial configurations \cite{Murphy1999MarsupiallikeMR}. These configurations consist of a team of robots, in which one acts as the provider robot, i.e. giving energy, transportation, communication, etc, to another robot. 
This paper follows this idea and considers the use of an UGV capable of deploying a tethered UAV to provide the latter with energy and high bandwidth communications through the cable, see Fig. \ref{fig:intro}.
This configuration poses significant challenges for trajectory planning
, 
as the shape of the cable should be taken into consideration to ensure collision-free navigation, especially in confined or underground environments, such as tunnels or sewers. We assume that there exists a device that allows us to control the length of the tether within a given range. We use this degree of freedom to prevent the tether from colliding with objects in the environment.

This paper proposes a method to generate optimal trajectories that allow to safely deploy a tethered UAV in such environments, preventing it and the tether from colliding against obstacles. The main contribution of the paper is the development of a formulation that considers a realistic shape of the tether that links the UAV and the UGV into the motion planning problem, in order to ensure the safety of the system. We validate the approach with simulation experiments in challenging realistic environments.

The rest of the paper is organized as follows. Section \ref{sec:related_work} analyzes the existing works in the literature that address similar problems. Then, Sections \ref{sec: approach} and \ref{sec:initilaization} describe the proposed methods for trajectory planning and trajectory initialization. The experimental results are discussed in Section \ref{sec:experimental}. Finally, the paper is concluded in Section \ref{sec:conclusions}, where we summarize the contributions of the paper and give insights of future research directions.

\section{Related Work}
\label{sec:related_work}

Global motion planning studies the problem of obtaining a feasible trajectory that allows a system to evolve from an starting configuration to 
a goal configuration \cite{LaValle}. Most methods assume a graph or grid that models the connectivity of the environment and then use \revision{graph search algorithms like} Djikstra, A* and Theta* \cite{NashKT10} to get an optimal path. To generate the graph, randomized methods such as PRM, RRT and its optimal variant RRT* \cite{karaman_rrt_star} have been successfully used in high-dimensional spaces. 

On the other hand, local planners aim to generate feasible trajectories in real-time parting from a global plan generated by one of the previous methods. They take into account the constraints in the motion of the particular platform and can handle the presence of unexpected obstacles. For instance, \cite{6698833} uses sparse graph optimization for obtaining optimal trajectories in UGVs, considering kinodynamic constraints and unmodeled obstacles. In this paper, we use a similar approach for generating optimal trajectories from a given initial path.

Regarding planning in marsupial systems, \cite{MooreIROS2018} proposes an interesting approach that uses a topological map that distinguishes the areas that are reachable by each robot in the system. The idea is to make profit of the marsupial configuration for reaching areas that would not be reachable with a single robot due to its size or autonomy. However, it does not consider tethers linking the different robots.

The research on aerial robots linked with a tether usually focuses on control aspects. \cite{6907304} proposes the use of a tether to increase the safety in landing maneuvers. Then, \cite{schulz_iros_18} considers it also to increase the stability when a UAV flights in confined spaces and \cite{8848946} analyzes the performance of several flight primitives for a tethered UAV. On the other hand, research on the use of tether is more mature in Unmanned Underwater Vehicles (UUVs) \cite{LARANJEIRA2020107018}, as the tether is used as a safety recovery device and as a communication link. However, it is usually assumed that the tethers move in a free space and thus collisions due to tethers are not considered. 

Regarding motion planning in marsupial tethered systems, \cite{6961531} presents a hierarchical approach to solve the path planning problem in map generation missions. It uses two independent RRT* for the UAV and UGV. In the first place, UGV plans its route and then the UAV limits its range according to the UGV plan and the tether length, but it does not consider the collisions due to the tether. In \cite{XiaoSSRR2018} the authors propose a relative localization system for a marsupial tethered UGV-UAV pair. They use the estimation of the angles, as measured by the cable's provider SDK and a catenary model to estimate the relative position of the UAV w.r.t. the UGV. Then, the same authors in \cite{XiaoIROS2018} propose a PRM planner in which reachable space is constrained by the tether in two ways: avoiding any contacts of the tether with obstacles, and allowing up to two different contact points to extend the effective range. However, experimental results showed that including contact points had an important negative impact in the relative localization of the platform, as the angles measured by the SDK are no longer a good estimation for the shape of the tether. Furthermore, the planner considers a tensed tether at any time, enabling collision checking based on ray-tracing. In contrast, we consider a hanging tether whose length is an additional configuration variable. The tether is thus not ensured to be tight and we have to model its shape as a catenary curve \cite{BOOKOFCURVES}.

\section{Approach}
\label{sec: approach}
This section describes the \revision{optimization-based method}. 
It is assumed that a map is available in the form of 3D occupancy grid. The goal is to generate a feasible trajectory for a UAV tethered to a UGV from the starting point (located on the UGV) to a goal position in the free space, considering dynamic constraints such as limited UAV velocities and accelerations, and including the length of the tether as a control variable. The UGV is assumed to remain static during the maneuver. The trajectory optimizer makes use of an existing global path as initial solution. This path is computed using the Lazy Theta* algorithm \cite{NashKT10} (see Section \ref{sec:initilaization} for details). 



\subsection{General Statement}

Our state space is defined by the UGV position $\mathbf{p}_g=(x_g,y_g,z_g)$, the UAV position $\mathbf{p}=(x,y,z)$ \revision{and the tether length $l$}. At any instant, the tether length $l \geq \|\mathbf{p} - \mathbf{p}_g\|$ and the positions of the UAV and UGV define the whole tether position through the catenary model \cite{BOOKOFCURVES}, whose parameters can be computed using the Bisection Numerical Method \cite{Numericalanalysis}. 

The objective is to determine the trajectory of the UAV $\mathbf{p}(t)$ and the length of the tether $l(t)$ (the UGV position will be considered as fixed) so that the UAV reaches a desired configuration while attaining a set of constraints posed by the obstacles, the kinematic and dynamic capabilities of the vehicles
. In our approach, such constraints will be encoded as a set of cost functions to minimize.

\subsection{Casting the problem as a sparse non-linear optimization}

In this work, we consider that the UGV position is fixed, so that the optimization problem reduces to estimating the UAV and tether length trajectories on time. We furthermore discretize the trajectory, so that the states of our problem become the set:

\begin{equation}
\label{eq:traj_params}
    O = \{\mathbf{p}_{i},l_i,\Delta t_{i}\}_{i=1,...,n}
\end{equation}

\noindent where $\Delta t_{i} = t_i - t_{i-1}$ is the time increment between consecutive states, allowing to consider the temporal aspects. Without loss of generality, we consider the UGV to be located at the origin of coordinates. The UAV takes-off from the UGV, so $\mathbf{p}_{0}=\mathbf{0}$ and $l_{0}=0$.

For each $\mathbf{p}_{i}$ and $l_i$, and considering that the UGV position is fixed and known, there is a tether configuration $TE_i$, given by the catenary model mentioned above. We also discretize the resultant tether model into a set of $m$ positions $\mathbf{p}_{te}=(x_{te}, y_{te}, z_{te})$:

\begin{equation}
    TE_i = \{\mathbf{p}_{te,j}\}_{j=0,...,m-1}
\end{equation}

Based on the discretization of the states, our problem consists of determining the values of the variables in $O$ that optimize a weighted multi-objective function $f(O)$:  
\begin{equation}
    O^* = \arg \min_O f(O) = \arg \min_O \sum_{k} \gamma_k * f_k(O) 
\end{equation}

\begin{figure}[!t]
    \centering
    \includegraphics[width=0.9\linewidth]{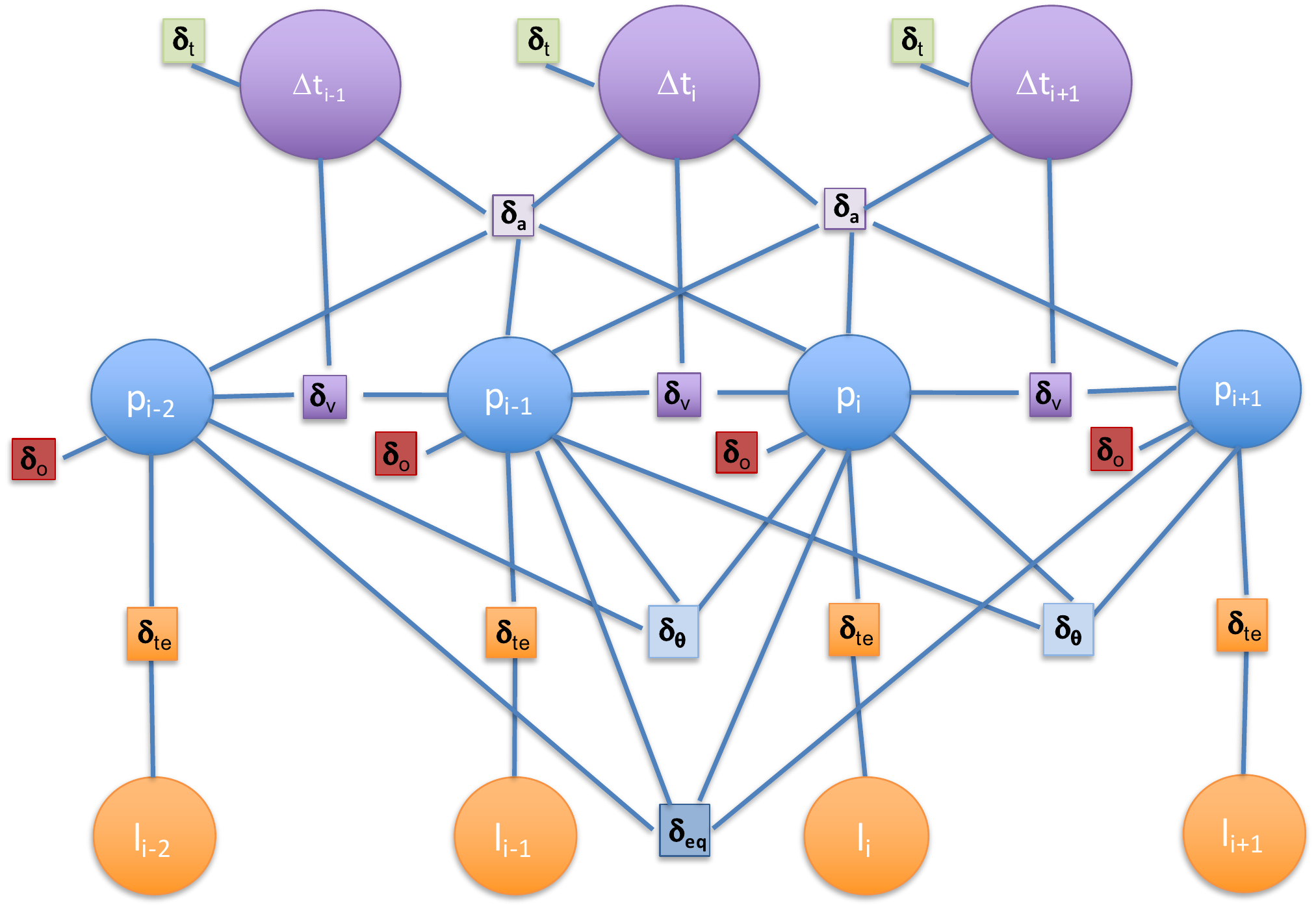}
    \caption{The optimization problem can be represented as a factor graph. Circles represent optimization variables $\mathbf{p}_{i},l_i,\Delta t_{i}$, and squares, factors (the different components of the cost function). The figure represent the factors for 4 consecutive positions (the links to factors and variables out from this time slice are not shown).}
    \label{fig:factor-graph}
    \vspace*{-5mm}
\end{figure}

\noindent $O^*$ denotes the optimized collision-free path for UAV and tether from start to goal point. \revision{The different cost terms are of the form $f_k(O)=\delta_{k}(O)^T\Omega_{k}\delta_{k}(O)$, that is, they are quadratic functions of non-linear \emph{factors} $\delta_{k}(O)$}. $\gamma_k$ is the weight for each component $f_k(O)$ of the objective. Each component encodes a different constraint or optimization objective of our problem, and will be presented next. All factors of the cost function are local with respect to $O$, as they only depend on a few number of consecutive configurations. Consequently, our optimization problem can be represented by a sparse factor graph (see Fig. \ref{fig:factor-graph}) and solved with non-linear sparse optimization algorithms. In particular, we use \emph{g$^2$o} \cite{KummerleICRA2011} as our optimization back-end.



\subsection{Constraints and Objective Function}

We introduce constraints in the graph generation such as tether length, velocity and acceleration limits and equidistance among consecutive robot poses. These constraints are introduced as penalty costs in the objective function. In addition, we aim to maximize the distance to obstacles and to minimize the length and execution time of the trajectory.

\subsubsection{Equidistance among vertices trajectory}
\label{sec:equi_constraints}
The optimization process might move vertices within the planning space in order to minimize the costs of the different constraints. To prevent grouping vertices in low-cost areas, we force them to keep a given distance with respect to previous and subsequent vertices. 

This is a multi-edge constraint that relates four consecutive UAV position vertices. It calculates the distance between each of the pairs of consecutive vertices and then computes the arithmetic average distance $\alpha$ \revision{among them}. The error $\epsilon$ is estimated between $\alpha$ and each distance between vertices.
\begin{equation}
    \delta_{eq,i} =  \begin{bmatrix} \alpha - \|p_{i-2}-p_{i-1}\|\\\alpha - \|p_{i-1}-p_{i}\|\\\alpha - \|p_{i}-p_{i+1}\|\end{bmatrix}
\end{equation}

\subsubsection{UAV obstacle avoidance}

This factor penalizes that the trajectory gets closer than a safety distance $\rho_{a}$ to the obstacles. This is an unary edge related with every position vertex. The distance $d_{i,o}$ from a vertex $\mathbf{p}_i$ to the closest obstacle is computed by a KD-Tree search over the 3D obstacle point-cloud. If the distance $d_{i,o}$ is inferior than the safety distance $\rho_{a}$ a high error is computed, otherwise the error is zero.
\begin{equation}
    \delta_{o,i} = \left \{
  \begin{array}{cc}
  e^{\rho_{a}-\beta*d_{i,o}} & ,if\  \  d_{i,o} \ < \ \rho_{o}\ \\
  0  & ,otherwise
  \end{array}
  \right .
\end{equation}


\noindent where $\beta$ is a constant parameter. 
\revision{The error increases exponentially as the distance to the obstacle gets closer to $0$.  }





\subsubsection{UAV kinematics}
We also introduce a multi-edge that relates three consecutive UAV position vertices. The aim of this edge is to smooth the optimized trajectory, avoiding abrupt direction changes. Two vectors are calculated, the first between the first and second consecutive vertices, \revision{$i-1$ and $i$}, and the second between the second and third consecutive vertices, \revision{$i$ and $i+1$}. Then, \revision{if the angle between the two vectors $\arccos(\frac{(\mathbf{p}_{i}-\mathbf{p}_{i-1})^{T}(\mathbf{p}_{i+1}-\mathbf{p}_{i})}{\|\mathbf{p}_{i}\|\|\mathbf{p}_{+1}\|})$ is larger} than a bound $\rho_\theta$,
an error is calculated:

\begin{equation}
    \delta_{\theta,i} = 
  \frac{1}{(\mathbf{p}_{i}-\mathbf{p}_{i-1})^{T}(\mathbf{p}_{i+1}-\mathbf{p}_{i})} 
\end{equation}


\subsubsection{UAV trajectory duration time}
We include an unary-edge related to time vertex. 
The edge is in charge of minimizing the trajectory time. 
The error is computed taking into account the time between two consecutive positions of the initial path and the current time trajectory.
\begin{equation}
    \delta_{t,i} = \Delta t^{[0]}_{i}-\Delta t_{i}
\end{equation}

\noindent where $\Delta t^{[0]}_{i-1,i}$ corresponds to the initial time difference reference for vertices  $i-1$ and $i$ and $\Delta t_{i}$ is the current value of the time difference vertex.

\subsubsection{UAV velocity}
It corresponds to a multi-edge that relates two consecutive position vertices and one time vertex. It is in charge to keep constant the speed during the optimized trajectory. Then, the error corresponds to the subtraction of the current and \revision{the desired velocity $\rho_v$ (a parameter of the algorithm)}. 
\begin{equation}
    \delta_{v,i} = \frac{\|\mathbf{p}_{i+1}-\mathbf{p}_{i}\|}{\Delta {t}_{i+1}} - \rho_v
\end{equation}



\subsubsection{UAV acceleration}
It corresponds to a multi-edge that relates three consecutive UAV position vertices and two consecutive time vertices. It is in charge of keeping the module of linear acceleration close to zero \revision{(to minimize control efforts)} 
for the optimized trajectory. To calculate the error two values of velocity and time are necessary. 

\begin{equation}
    \delta_{a,i} = \frac{\|\mathbf{v}_{i}\|-\|\mathbf{v}_{i-1}\|}{\Delta {t}_{i} + \Delta {t}_{i+1}}
\end{equation}

\noindent where $\mathbf{v}_{i}=\frac{\mathbf{p}_{i+1}-\mathbf{p}_{i}}{\Delta {t}_{i+1}}$ is the velocity between consecutive poses.

\subsubsection{Tether obstacle avoidance}
This is a binary-edge that relates the UAV position vertex and the tether length vertex. Its main objective is to avoid the tether collision. Two error components are calculated. The first avoids positioning the tether in places where it collides. If the distance from tether points to obstacles $d_{i,to}$ is lower that a safety distance $\rho_{t}$ then the error is computed, otherwise, this component is zero. The error favors longer lengths of the tether. The distance to obstacles is computed by using kd-tree search for the points of the tether discretized model $TE_i$.

The second component forces the tether to stretch if it is free from collision, that is, when $d_{i,to}$ is greater than the security distance $\rho_{l}$.
\begin{eqnarray}
  \delta_{l,i}[0] =& \left \{
  \begin{array}{cc}
  10^{4} \cdot e^{10(\|\mathbf{p}_i\|-l_{i})} & ,if\  \  d_{i,to} \ < \ \rho_{l}\ \\\
  0  & ,otherwise
  \end{array}
  \right . \\
  \delta_{l,i}[1] =& \left \{
  \begin{array}{cc}
  0 & ,if\  \  d_{i,to} \ < \ \rho_{l}\ \\
  (l_{i} - \|\mathbf{p}_i\|) &, otherwise
  \end{array}
  \right . 
\end{eqnarray}

\noindent  where $\|\mathbf{p}_i\|$ is the distance between UAV and UGV and $l_{i}$ the tether length. The relative weight is much larger for the collision case as this is a hard constraint to prioritize tether safety.




\section{Problem initialization}
\label{sec:initilaization}

Non-linear optimization problems need of a proper initialization in order to reach the optimal solution. Using naive solutions as, for instance, a straight line between starting and goal positions, could make the optimizer converge into a local minimum.

We need to initialize the trajectory parameters defined in (\ref{eq:traj_params}) and also the number of states $n$. Ideally, the initial solution must be feasible, that is, the different states must be reachable by the UAV without colliding with obstacles (both the UAV and the tether). Unfortunately, there are no approaches in the state of the art dealing with this problem. Here, we have adapted the Lazy Theta* \cite{NashKT10} algorithm in order to properly consider the constraints derived from the tether connecting UAV and UGV. Once a feasible sequence of states is computed by the planning algorithm, time information is added to the computed path as initial solution for the trajectory. Next paragraphs describe these steps with detail.

\subsection{Path estimation}
As previously mentioned, a Lazy Theta* \cite{NashKT10} path planner has been adapted in order to include the constraints of the tether into the plan computation without expanding the dimension of the planning state in order to keep computation as low as possible.

Thus, the function that checks the feasibility of a given state has been updated to also consider the tether. Once a state is considered as feasible, a second check is performed to evaluate the tether at that state using the function described in Algorithm \ref{alg:catenary_check}. It starts with the minimal possible tether length and increases the length while the tether is in collision with an obstacle at any point. When no collision is detected and the tether does not reach the maximum length $l_{max}$, the state $\mathbf{p}$ is considered as feasible for the catenary. 

The function $isCatenaryInCollision(\mathbf{p},\mathbf{p}_g,l)$  in Algorithm \ref{alg:catenary_check} computes the catenary curve between $\mathbf{p}$ and the UGV configuration $\mathbf{p}_g$ subject to the tether length $l$, and discretizes it. For each point of the catenary, the function computes the distance to the closest obstacle in the planning space. If this distance if over a given threshold for all the points of the catenary, the configuraton $\{\mathbf{p}, \mathbf{p}_g, l\}$ is considered as free of collision.

\begin{algorithm}[!t]

\label{alg:catenary_check}
\SetKwProg{checkCatenaryFeasibility}{checkCatenaryFeasibility(p,p$_g$)}{}{end}

\SetAlgoLined
\checkCatenaryFeasibility{}{
 $l=||\mathbf{p}-\mathbf{p}_g||$\;
 \While{isCatenaryInCollision($\mathbf{p}$,$\mathbf{p}_g,l$) \&\& $l < l_{max}$}{
  $l=l+\epsilon$\;
 }
 \eIf{$l < l_{max}$}{
   return TRUE\;
   }{
   return FALSE\;
  }
  }
 \caption{Catenary feasibilty check between state $\mathbf{p}$ and UGV configuration $\mathbf{p}_g$. Return TRUE if $\mathbf{p}$ is feasible.}
\end{algorithm}

\begin{figure*}[ht]
    \centering
    \includegraphics[width=\textwidth]{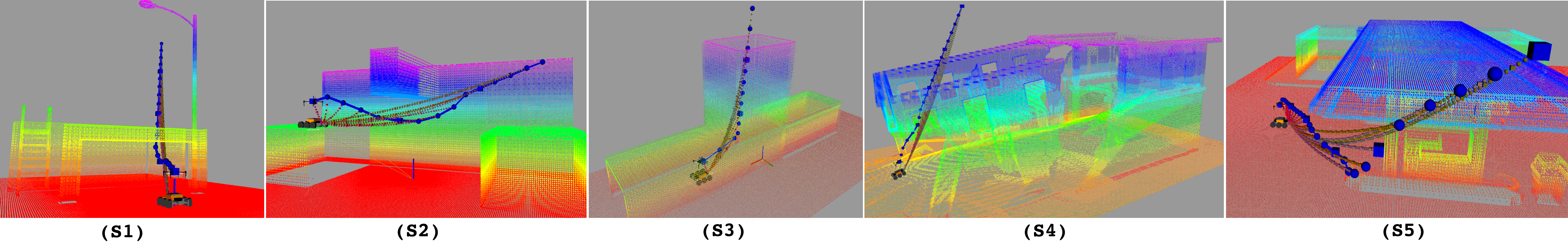}
    \caption{Scenarios considered for validation.(a) S1: Open space with arc as obstacle. (b) S2: Narrow/constrained space with denied access to UGV. (c) S3: Confined space with outlet duct for UAV. (d) S4: Collapsed Fire Station. (e) S5: Open space gas station.}
    \label{fig:exp_scenarios}
    \vspace*{-5mm}
\end{figure*}

\subsection{Trajectory estimation}
The optimal nature of the Lazy Theta* algorithms prevents the creation of way-points when there exists direct line of sight with respect to the previous way-point. As a result, the computed path tends to be sampled frequently in complex areas, while open areas are traversed with a straight line.  

However, we are interested in sampling the trajectory at regular intervals, so that kinematic constraints and obstacle costs can be better evaluated. Actually, one of the constraints imposed by the optimizer is the equidistance among trajectory way-points (Section \ref{sec:equi_constraints}).

Thus, the computed Lazy Theta* path is interpolated at fixed intervals between way-points, adding points in the middle to fulfil the equidistance constraint. Fortunately, the path planner guarantees that all points in the line that connects two way-points are feasible because they comply with the line-of-sight constraint of the Lazy Theta*. The length of the tether in the interpolated points are computed using Algorithm \ref{alg:catenary_check}.

Finally, this solution does not consider any kinematic constraints, it is just a collision-free path from the initial UAV position to the goal. Therefore, the values of $\Delta t_{i}$ in (\ref{eq:traj_params}) are not yet initialized. To do so, a constant velocity model is imposed over the path, so that $\Delta t_{i}$ can be computed from sequenced way-points considering the Euclidean distance between them.

The solution provided by this method is optimal in terms of the traversed distance by the UAV considering catenary constraints, but the catenary length is not optimal at each state, just feasible. This solution neither consider kinematic constraints on the trajectory or distance to obstacle. All these parameters will be computed/refined in the optimization process.

\section{Experimental Results}
\label{sec:experimental}
A set of experiments in simulated environments has been conceived to validate the feasibility of the proposed approach. While the algorithm has been tested on simulated environments, the full approach has been implemented as a planner in Robot Operating System (ROS).

The trajectory planner has been implemented in C++, and it is publicly available\footnote{\url{https://github.com/robotics-upo/marsupial_optimizer}, branch: g2o}. \emph{g$^2$o} \cite{KummerleICRA2011} has been selected as the engine for graph optimization. The solution makes use of classical approaches to solve computationally demanding tasks, such as kd-trees for distance to closest point computation.

The approach has been validated in five different scenarios (S) with different complexity, i.e. confined spaces (S3 and S4), open spaces (S1 and S5) and narrow/constrained spaces (scenario S2), shown in the snapshot from Figure \ref{fig:exp_scenarios}. The accompanying video summarizes each of five experiments in the different scenarios for better understanding. For each scenario, the robot has been placed in different initial positions and several goals have been sent to the planner. All the experiments were executed with the same set of parameters, setting the maximum number of iterations of the optimizer to 100 and the rest parameters to: 
\begin{itemize}
    \item Weighting factors (in brackets the associated constraint): $\gamma_{eq}$ = 0.6 (1), $\gamma_{o}$ = 0.8 (2), $\gamma_{\theta}$ = 0.4 (3), $\gamma_{t}$ = 0.005 (4), $\gamma_{v}$ = 0.06 (5), $\gamma_{a}$ = 1.0 (6), $\gamma_{l}$ = 0.9 (7). 
    \item Avoiding collision constr.:  $\beta$ = 4.0, $\rho_{o}$ = 1.0,  $\rho_{l}$ = 0.2. 
    \item Kinematic constraints: $\rho_\theta $ = $\frac{\pi}{6}$, $\rho_v$ = 2.0.
\end{itemize}
The results of the experiments are detailed in Tables \ref{tab:experiments} and \ref{tab:computation_time}. The metrics presented are the following:
\begin{itemize}
\item SIG: Scenario - Initial Position - Goal Position
\item LIP: Length of initial path [m]	
\item LTO: Length of the optimized trajectory [m] 	
\item TIP: Time of the initial path	[s]
\item TOT: Time of the optimized trajectory [s] 	
\item DOI: Distance obstacles-initial path [m]	
\item DOO: Distance obstacles-optimized trajectory [m]
\item DCOI: Dist. catenary-obstacle solution initial [m]
\item DCOO: Dist. catenary-obstacles solution optimized[m]	
\item VTO: Velocity of the optimized trajectory [m/s] 	
\item ATO: Acceleration of the optimized trajectory [m/s$^2$]
\item TCI: Time compute initial solution
\item TCO: Time compute optimized solution
\end{itemize}

In order to benchmark the solution, we use as baseline the value of the different metrics prior to optimization, that is, the solution of the interpolated Lazy Theta* detailed in Section \ref{sec:initilaization}. Next paragraphs evaluate the different metrics.

\begin{table*}[ht]
\caption{Results for experiments in simulated environments.}
\centering
\label{tab:experiments}
\begin{adjustbox}{max width=\textwidth}
\begin{tabular}{|l|ll|ll|llll|llll|llll|}
\hline
\multicolumn{1}{c}{SIG} & \multicolumn{1}{c}{LIP} & \multicolumn{1}{c}{LTO} & \multicolumn{1}{c}{TIP} & \multicolumn{1}{c}{TOT} & \multicolumn{1}{c}{\begin{tabular}[c]{@{}c@{}}Mean\\ DOI\end{tabular}} & \multicolumn{1}{c}{\begin{tabular}[c]{@{}c@{}}Min\\ DOI\end{tabular}} & \multicolumn{1}{c}{\begin{tabular}[c]{@{}c@{}}Mean\\ DOO\end{tabular}} & \multicolumn{1}{c}{\begin{tabular}[c]{@{}c@{}}Min\\ DOO\end{tabular}} & \multicolumn{1}{c}{\begin{tabular}[c]{@{}c@{}}Mean\\ DCOI\end{tabular}} & \multicolumn{1}{c}{\begin{tabular}[c]{@{}c@{}}Min\\ DCOl\end{tabular}} & \multicolumn{1}{c}{\begin{tabular}[c]{@{}c@{}}Mean\\ DCOO\end{tabular}} & \multicolumn{1}{c}{\begin{tabular}[c]{@{}c@{}}Min\\ DCOO\end{tabular}} & \multicolumn{1}{c}{\begin{tabular}[c]{@{}c@{}}Mean\\ VTO\end{tabular}} & \multicolumn{1}{c}{\begin{tabular}[c]{@{}c@{}}Max\\ VTO\end{tabular}} & \multicolumn{1}{c}{\begin{tabular}[c]{@{}c@{}}Mean\\ ATO\end{tabular}} & \multicolumn{1}{c}{\begin{tabular}[c]{@{}c@{}}Max\\ ATO\end{tabular}} \\
\hhline{|=|=|=|=|=|=|=|=|=|=|=|=|=|=|=|=|=|}
S1.1.1 & 3,97 & 3,98 & 1,99 & 1,99 & 0,84 & 0,51 & 0,84 & 0,54 & 0,70 & 0,28 & 0,69 & 0,28 & 2,0030 & 2,0120 & 0,0010 & 0,0140 \\
S1.1.2 & 4,95 & 5,22 & 2,48 & 2,54 & 0,88 & 0,51 & 0,97 & 0,83 & 0,65 & 0,21 & 0,35 & 0,17 & 2,0550 & 2,0630 & 0,0010 & 0,0070 \\
S1.1.3 & 5,34 & 5,89 & 2,67 & 2,81 & 1,01 & 0,51 & 1,18 & 0,90 & 0,72 & 0,23 & 0,69 & 0,23 & 2,0930 & 2,0960 & -0,0010 & 0,0010 \\
S1.1.4 & 6,27 & 6,77 & 3,13 & 3,30 & 0,91 & 0,49 & 1,13 & 0,78 & 0,81 & 0,44 & 0,80 & 0,33 & 2,0500 & 2,0510 & 0,0000 & 0,0000 \\
S1.1.5 & 10,00 & 10,13 & 5,00 & 5,03 & 1,55 & 0,48 & 1,59 & 0,73 & 1,00 & 0,24 & 0,45 & 0,22 & 2,0160 & 2,0280 & -0,0030 & 0,0100 \\
S1.1.6 & 11,59 & 11,73 & 5,80 & 5,83 & 1,35 & 0,49 & 1,39 & 0,73 & 1,04 & 0,21 & 0,36 & 0,22 & 2,0140 & 2,0290 & -0,0030 & 0,0040 \\
S1.2.1 & 7,03 & 7,67 & 3,51 & 3,68 & 0,96 & 0,43 & 1,12 & 0,88 & 0,71 & 0,21 & 0,58 & 0,08 & 2,0830 & 2,0870 & 0,0010 & 0,0020 \\
S1.2.3 & 8,38 & 9,03 & 4,19 & 4,33 & 1,04 & 0,43 & 1,21 & 0,93 & 0,69 & 0,18 & 0,60 & 0,07 & 2,0860 & 2,1270 & 0,0080 & 0,0240 \\
S1.2.5 & 9,75 & 9,85 & 4,88 & 5,01 & 2,40 & 1,37 & 2,44 & 1,71 & 1,47 & 0,56 & 1,51 & 0,56 & 1,9650 & 1,9650 & 0,0000 & 0,0000 \\
S1.2.6 & 11,69 & 11,82 & 5,84 & 6,00 & 2,01 & 1,54 & 2,10 & 1,65 & 1,58 & 0,58 & 1,58 & 0,58 & 1,9700 & 1,9710 & 0,0000 & 0,0000 \\
\hline
S2.1.1 & 8,93 & 9,64 & 4,47 & 4,67 & 0,80 & 0,38 & 1,15 & 0,88 & 0,68 & 0,23 & 0,79 & 0,23 & 2,0630 & 2,0670 & -0,0010 & 0,0000 \\
S2.1.2 & 9,00 & 9,58 & 4,50 & 4,76 & 1,10 & 0,38 & 1,31 & 0,88 & 0,83 & 0,21 & 0,89 & 0,21 & 2,0130 & 2,0170 & -0,0010 & 0,0010 \\
S2.1.3 & 9,25 & 9,72 & 4,63 & 4,72 & 1,26 & 0,38 & 1,41 & 0,88 & 0,93 & 0,22 & 0,99 & 0,22 & 2,0580 & 2,0630 & -0,0010 & 0,0000 \\
S2.1.4 & 5,91 & 6,76 & 2,96 & 3,24 & 0,72 & 0,39 & 1,02 & 0,73 & 0,70 & 0,20 & 0,79 & 0,21 & 2,0830 & 2,0840 & 0,0000 & 0,0010 \\
S2.1.5 & 6,00 & 7,22 & 3,00 & 3,45 & 0,79 & 0,38 & 1,16 & 0,88 & 0,81 & 0,21 & 0,92 & 0,21 & 2,0960 & 2,0980 & -0,0010 & 0,0010 \\
S2.1.6 & 6,47 & 7,39 & 3,24 & 3,54 & 0,79 & 0,38 & 1,09 & 0,88 & 0,89 & 0,21 & 1,92 & 0,20 & 2,0850 & 2,1490 & 0,0190 & 0,0440 \\
S2.2.1 & 9,00 & 9,62 & 4,50 & 4,77 & 0,92 & 0,38 & 1,14 & 0,90 & 0,86 & 0,21 & 0,95 & 0,21 & 2,0180 & 2,0290 & -0,0030 & 0,0000 \\
S2.2.2 & 8,93 & 9,35 & 4,47 & 4,63 & 1,18 & 0,38 & 1,34 & 0,95 & 0,95 & 0,23 & 1,01 & 0,23 & 2,0210 & 2,0260 & -0,0020 & 0,0000 \\
S2.2.3 & 9,06 & 9,65 & 4,53 & 4,72 & 1,25 & 0,38 & 1,43 & 0,97 & 0,93 & 0,20 & 1,01 & 0,20 & 2,0420 & 2,0480 & -0,0010 & 0,0000 \\
S2.2.4 & 6,00 & 7,09 & 3,00 & 3,39 & 0,82 & 0,38 & 1,14 & 0,88 & 0,89 & 0,20 & 1,01 & 0,21 & 2,0900 & 2,0960 & 0,0010 & 0,0030 \\
S2.2.5 & 5,97 & 6,97 & 2,99 & 3,30 & 0,83 & 0,38 & 1,13 & 0,95 & 0,90 & 0,21 & 1,01 & 0,20 & 2,1100 & 2,1260 & 0,0040 & 0,0130 \\
S2.2.6 & 6,13 & 6,58 & 3,06 & 3,17 & 0,83 & 0,38 & 0,98 & 0,84 & 0,90 & 0,21 & 1,04 & 0,20 & 2,0790 & 2,0910 & 0,0030 & 0,0140 \\
S2.3.1 & 9,25 & 9,73 & 4,63 & 4,72 & 1,04 & 0,38 & 1,19 & 0,90 & 0,93 & 0,22 & 0,99 & 0,22 & 2,0600 & 2,0740 & -0,0040 & 0,0000 \\
S2.3.2 & 9,06 & 9,65 & 4,53 & 4,73 & 1,21 & 0,38 & 1,39 & 0,97 & 0,93 & 0,20 & 1,01 & 0,20 & 2,0420 & 2,0480 & -0,0010 & 0,0030 \\
S2.3.3 & 8,93 & 9,34 & 4,47 & 4,63 & 1,22 & 0,38 & 1,39 & 0,94 & 0,95 & 0,23 & 1,01 & 0,23 & 2,0190 & 2,0240 & -0,0020 & 0,0000 \\
S2.3.4 & 6,40 & 7,34 & 3,20 & 3,49 & 0,82 & 0,38 & 1,11 & 0,88 & 0,90 & 0,21 & 1,01 & 0,21 & 2,1010 & 2,1040 & 0,0000 & 0,0020 \\
\hline
S3.1.1 & 8,00 & 8,00 & 4,00 & 4,00 & 0,91 & 0,86 & 0,91 & 0,86 & 0,50 & 0,21 & 0,49 & 0,21 & 2,0000 & 2,0050 & 0,0000 & 0,0170 \\
S3.1.2 & 12,60 & 12,60 & 6,30 & 6,30 & 0,91 & 0,86 & 0,92 & 0,86 & 0,44 & 0,14 & 0,43 & 0,12 & 2,0000 & 2,0050 & 0,0010 & 0,0110 \\
S3.2.1 & 5,40 & 5,40 & 2,70 & 2,70 & 0,90 & 0,84 & 0,91 & 0,86 & 0,54 & 0,25 & 0,52 & 0,21 & 1,9990 & 2,0030 & 0,0010 & 0,0150 \\
S3.2.2 & 10,00 & 10,00 & 5,00 & 5,00 & 0,91 & 0,84 & 0,92 & 0,85 & 0,47 & 0,18 & 0,47 & 0,18 & 2,0000 & 2,0030 & 0,0000 & 0,0140 \\
S3.2.3 & 5,63 & 5,64 & 2,82 & 2,82 & 0,89 & 0,61 & 0,89 & 0,62 & 0,70 & 0,26 & 0,69 & 0,25 & 2,0000 & 2,0030 & 0,0000 & 0,0080 \\
S3.2.4 & 7,81 & 7,82 & 3,91 & 3,91 & 0,97 & 0,61 & 0,98 & 0,63 & 0,67 & 0,20 & 0,66 & 0,20 & 2,0000 & 2,0030 & 0,0000 & 0,0080 \\
\hline
S4.1.1 & 7,83 & 7,83 & 3,91 & 3,91 & 1,57 & 0,89 & 1,57 & 0,89 & 1,19 & 0,55 & 1,20 & 0,55 & 2,0000 & 2,0010 & 0,0000 & 0,0060 \\
S4.1.2 & 11,42 & 11,57 & 5,71 & 5,73 & 2,23 & 0,89 & 2,29 & 0,89 & 1,64 & 0,55 & 1,64 & 0,56 & 2,0200 & 2,0200 & 0,0000 & 0,0000 \\
\hline
S5.1.1 & 10,32 & 10,33 & 5,16 & 5,16 & 1,31 & 0,85 & 1,31 & 0,85 & 1,02 & 0,38 & 1,02 & 0,35 & 2,0010 & 2,0030 & 0,0000 & 0,0010 \\
S5.1.2 & 10,41 & 10,63 & 5,21 & 5,34 & 1,52 & 1,02 & 1,70 & 1,28 & 1,37 & 0,49 & 1,44 & 0,48 & 1,9910 & 1,9910 & 0,0000 & 0,0000 \\
S5.1.3 & 14,20 & 14,51 & 7,10 & 7,22 & 1,40 & 0,44 & 1,51 & 0,98 & 1,09 & 0,32 & 1,06 & 0,32 & 2,0110 & 2,0130 & 0,0000 & 0,0010 \\
S5.1.4 & 13,94 & 14,51 & 6,97 & 7,19 & 1,36 & 0,45 & 1,59 & 0,98 & 1,00 & 0,23 & 0,99 & 0,17 & 2,0160 & 2,0210 & 0,0000 & 0,0020 \\
\hline
\end{tabular}
\end{adjustbox}
\vspace*{-2mm}
\end{table*}
\begin{table*}[!ht]
\caption{Computational time in each of the simulated scenarios}
\label{tab:computation_time}
\centering
\begin{tabular}{|l||lllllllllllll|}
\hline
SIG & 1.1.1 & 1.1.2 & 1.1.3 & 1.1.4 & 1.1.5 & 1.1.6 & 1.2.1 & 1.2.3 & 1.2.5 & 1.2.6 & 2.1.1 & 2.1.2 & 2.1.3 \\
\hhline{|=|=|=|=|=|=|=|=|=|=|=|=|=|=|}
\begin{tabular}[c]{@{}l@{}}Mean TCI\end{tabular} & 0,84 & 24,38 & 2,64 & 1,02 & 32,83 & 40,64 & 7,66 & 21,80 & 0,48 & 0,64 & 0,51 & 0,62 & 0,70 \\
\begin{tabular}[c]{@{}l@{}}Mean TCO\end{tabular} & 1,35 & 0,99 & 2,38 & 3,91 & 4,49 & 10,10 & 1,47 & 4,98 & 9,71 & 14,97 & 9,21 & 12,16 & 10,14 \\
\hline
SIG & 2.1.4 & 2.1.5 & 2.1.6 & 2.2.1 & 2.2.2 & 2.2.3 & 2.2.4 & 2.2.5 & 2.2.6 & 2.3.1 & 2.3.2 & 2.3.3 & 2.3.4 \\
\hhline{|=|=|=|=|=|=|=|=|=|=|=|=|=|=|}
\begin{tabular}[c]{@{}l@{}}Mean TCI\end{tabular} & 41,10 & 24,48 & 30,18 & 0,69 & 0,67 & 0,77 & 21,26 & 36,72 & 36,75 & 0,74 & 0,76 & 0,69 & 49,48 \\
\begin{tabular}[c]{@{}l@{}}Mean TCO\end{tabular} & 4,83 & 5,21 & 7,70 & 12,12 & 12,55 & 14,47 & 5,06 & 7,32 & 5,18 & 11,50 & 12,99 & 10,14 & 7,17 \\
\hline
SIG & 3.1.1 & 3.1.2 & 3.2.1 & 3.2.2 & 3.2.3 & 3.2.4 & 4.1.1 & 4.1.2 & 5.1.1 & 5.1.2 & 5.1.3 & 5.1.4 &  \\
\hhline{|=|=|=|=|=|=|=|=|=|=|=|=|=|=|}
\begin{tabular}[c]{@{}l@{}}Mean TCI\end{tabular} & 0,37 & 0,38 & 0,38 & 0,38 & 0,58 & 1,21 & 8,15 & 1,16 & 0,46 & 0,84 & 1,42 & 49,07 &  \\
\begin{tabular}[c]{@{}l@{}}Mean TCO\end{tabular} & 4,36 & 7,85 & 2,41 & 6,15 & 4,56 & 7,71 & 10,75 & 21,30 & 12,61 & 18,95 & 25,34 & 25,12 & \\
\hline
\end{tabular}
\vspace*{-5mm}
\end{table*}

\subsection{Trajectory length and elapsed time}

The initial trajectories are generated by our Lazy Theta* algorithm, which is focused on minimizing the length of the path and, assuming constant speed, the time spent. Thus, the lengths of the optimized trajectories cannot outperform the initial ones in these terms, and sometimes there are noticeable differences as it considers another objectives such as obstacle clearance and kinodynamic constraints.  

The good news are that the differences rarely exceed 10\% of the optimal length and that the generated trajectories are  smoother and 
\revision{safer than initial path}.

\subsection{Distance to obstacles}
The simulation results \revision{(DOI, DOO, DCOI, DCOO)} confirm how the optimized trajectory tends to move away from obstacles when the trajectory itself or the tether is closer than the safety distance. 
As seen in the previous section, this comes at a cost. To increase the clearance to the obstacles, the length of the path should also increase with respect to the initial path.

\subsection{Trajectory kinematics}
\begin{figure}[!t]
    \centering
    \includegraphics[width=0.8\linewidth]{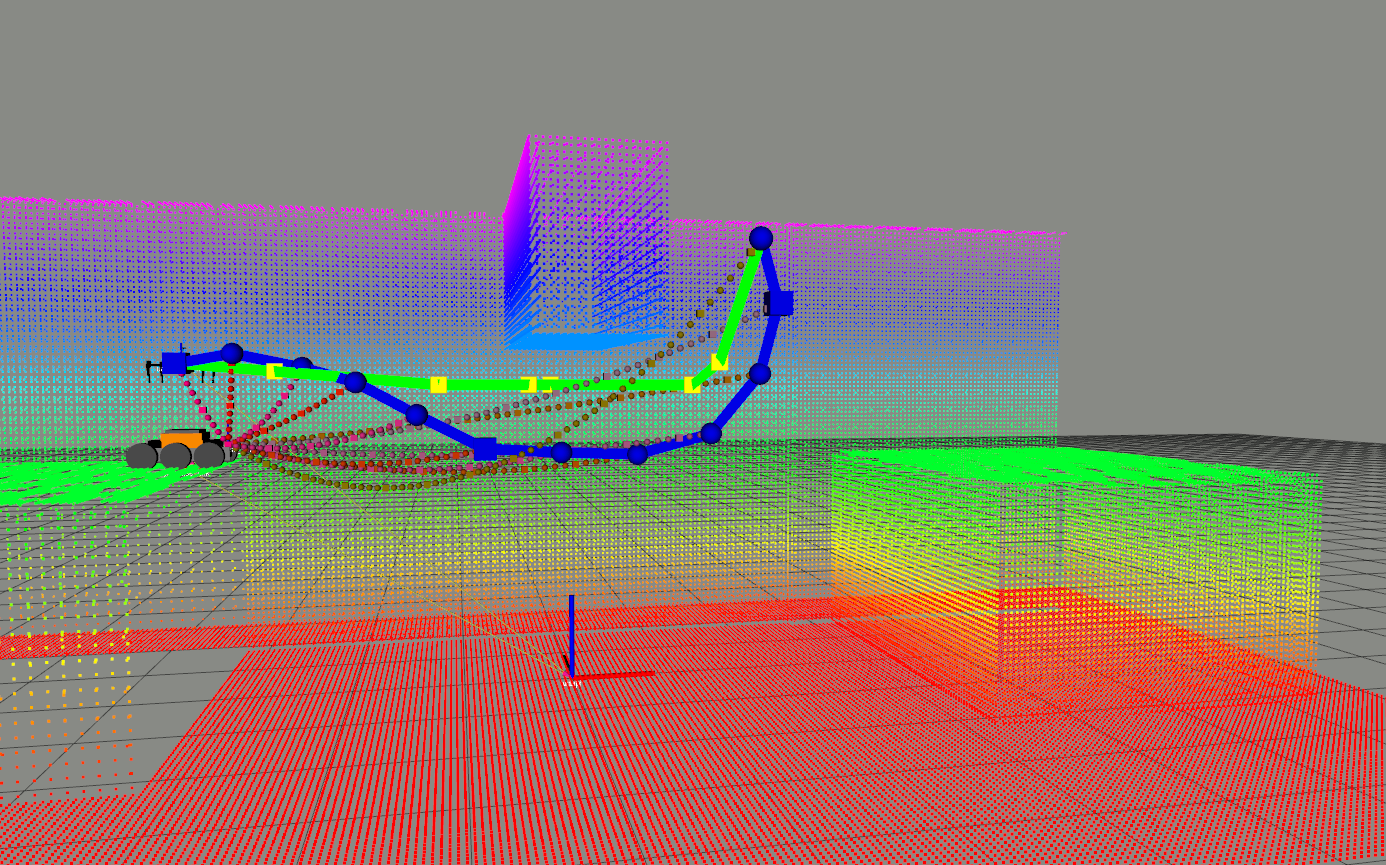}
    \caption{Show how the optimized trajectory (blue line) is smoother than initial path (green line). Tether configuration is shown as dotted curved between planned way-point and UGV.}
    \label{fig:kinematics}
    \vspace*{-5mm}
\end{figure}

The kinematic constraints allows us to obtain a smoother trajectory with respect to the initial path. \revision{Figure \ref{fig:kinematics} shows an example in which the benefits of the trajectory smoothing (blue line) can be clearly seen in comparison to the initial path (green line)}. The results in Table \ref{tab:experiments} also show how the velocity and acceleration \revision{(VTO, ATO)} are close to the desired values (2.0 $m/s$ and 0 $m/s^2$ respectively).

\subsection{Computation time}

Table \ref{tab:computation_time} presents a summary of the computational time required to estimate both the initial and optimized trajectories. We used a Intel Core i7 2.20GHz with 16 GB of RAM. The results indicate that computation times of the Lazy Theta* algorithm presents a great variability, from fractions of second to 40 seconds, even when dealing with problems in the same scenario. On the other hand, the optimization algorithm gives results ranging from one to twenty-five seconds. As a general tendency, we have verified that the computation time growths with larger trajectories and with increasing clutter. Nevertheless, we consider this timing is acceptable for this type of complex planning problems.

\section{Conclusions and Future Work}
\label{sec:conclusions}
This paper presented a method for trajectory planning of a tethered UAV based on optimization. To the best of our knowledge, this is the first planning approach of this kind for this type of problem that considers the curvature of a non-tight tether. The method has been intensively tested on simulated scenarios and the results are promising as the proposed approach is able to optimize the UAV trajectory together with its kinematics constraints taking into account the limitations of the tether. 
Solutions are obtained quickly, in a matter of seconds in most cases, in a regular computer. 

Unfortunately, this solution does not consider the movement of the UGV to compute trajectories of the whole system. Thus, future works will consider extending this work to include the UGV as a moving system (marsupial system) which trajectory can be also optimized together with the UAV and the tether, as well as the tether deployment dynamics.



\balance
\bibliographystyle{IEEEtran} 
\bibliography{IEEEabrv}

\end{document}